# Hubiness, length, crossings and their relationships in dependency trees

*Ramon Ferrer-i-Cancho[1]*

**Abstract.** Here tree dependency structures are studied from three different perspectives: their degree variance (hubiness), the mean dependency length and the number of dependency crossings. Bounds that reveal pairwise dependencies among these three metrics are derived. Hubiness (the variance of degrees) plays a central role: the mean dependency length is bounded below by hubiness while the number of crossings is bounded above by hubiness. Our findings suggest that the online memory cost of a sentence might be determined not just by the ordering of words but also by the hubiness of the underlying structure. The 2$^{nd}$ moment of degree plays a crucial role that is reminiscent of its role in large complex networks.

*Keywords: syntactic dependency, syntax, dependency length, crossings.*

## 1. Introduction

According to dependency grammar (Mel'čuk 1988, Hudson 2007) the structure of a sentence can be defined by means of a tree in which arcs indicate syntactic dependencies between the occurrences of words (Fig. 1). In standard graph theory (Bollobás 1998), the black circles from which arcs arrive or depart in Fig. 1 (black circles) are called vertices. Vertices are usually labeled with words. Thus, each occurrence of a word of a sentence corresponds to a vertex. Arcs are also called edges or links. Here we focus on two aspects of dependency trees: the length of the dependencies (the distance between syntactically linked words) and the number of crossings of the dependency tree. The syntactic dependency structure of a sentence (Fig. 1) is perhaps the most inspiring and useful linguistic example of dependency tree. This article is motivated by those trees.

We assume that the words of a sentence are placed in a sequence in the same order as in the original sentence and define the concept of distance in this sequence. We adopt the convention that the position of the first word of the sentence (i.e. the 1$^{st}$ element of the sequence) is 1, the position of the second word of the sentence (i.e. the 2$^{nd}$ element of the sequence) is 2 and so on. $\pi(v)$ is defined as the position of a vertex $v$. In Fig. 1, $\pi(\text{'she'}) = 1$, $\pi(\text{'loved'}) = 2$ and so on. $n$ is defined as the length of the sentence in words. $n$ is also the number of vertices of the tree and the position of the last word of the sentence. *d* is defined as the distance between two vertices *u* and *v* as the absolute difference of their positions, i.e. $d = |\pi(u) - \pi(v)|$. If *u* and *v* are linked, then *d* is also the length of the edge formed by vertices *u* and *v* (Ferrer-i-Cancho 2004). Thus the distance or the length of the dependency between '*she*' and '*loved*' is $d = 1$ and the distance or the length of the dependency between '*loved*' and '*for*' is $d = 2$. *d* goes from 1 to $n - 1$.

---

[1] Complexity and Quantitative Linguistics Lab. Departament de Llenguatges i Sistemes Informàtics, TALP Research Center, Universitat Politècnica de Catalunya (UPC). Campus Nord, Edifici Ω, Jordi Girona Salgado 1-3. 08034 Barcelona, Catalonia (Spain). Phone: +34 934137870, Fax: +34 934137787. E-mail: rferrericancho@lsi.upc.edu

Alternatively, dependency length has been defined so that consecutive words have distance zero (e.g. Hudson 1995, Hiranuma 1999). $d_0$ is used for referring to the length or distance defined using this alternative convention. This way, the length of the dependency between 'she' and 'loved' is $d_0=0$ and that of the dependency between 'loved' and 'for' is $d_0=1$. $d_0$ goes from 0 to $n-2$.

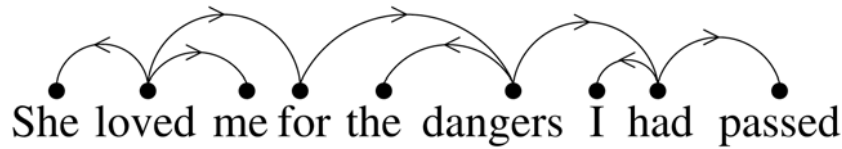

Figure 1. The syntactic structure of the sentence *'She loved me for the dangers I had passed'* following the conventions by Mel'čuk (1988). *'she'* and the verb *'loved'* are linked by a syntactic dependency. Arcs go from governors to dependents. Thus, '*she*' and '*me*' are dependents of the verbal form '*loved*'. Indeed, *'she'* and *'me'* are arguments of the verb form *'loved'* (the former as subject and the latter as object).

The concept of link crossing (Hays 1964, Holan et al. 2000, Hudson 2000, Havelka 2007) will be defined next. Imagine that we have two pairs of linked vertices: $(u,v)$ and $(x,y)$, such that $\pi(u) < \pi(v)$ and $\pi(x) < \pi(y)$. The arcs (or edges) of $(u,v)$ and $(x,y)$ cross if and only if $\pi(u) < \pi(x) < \pi(v) < \pi(y)$ or $\pi(x) < \pi(u) < \pi(y) < \pi(v)$. We define *C* as the number of different pairs of edges that cross. For instance, $C = 0$ in the sentence in Fig. 1 and $C = 9$ in Fig. 2. When there are no vertex crossings ($C = 0$), the syntactic dependency tree of a sentence is said to be planar (Havelka 2007).

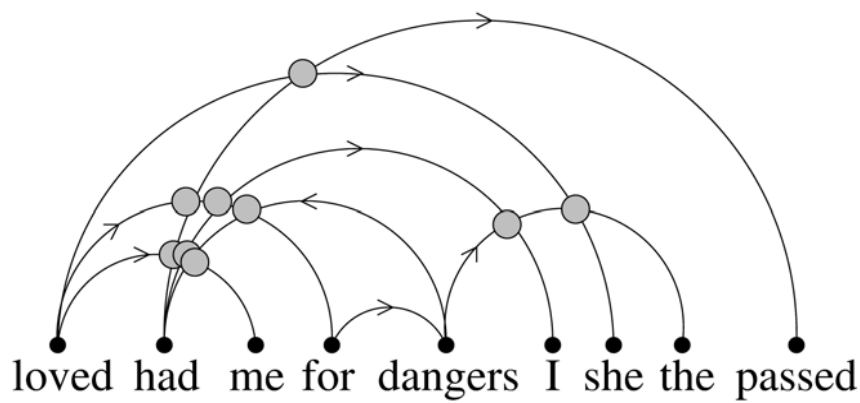

Figure 2. The structure of the sentence in Fig. 1 after having scrambled the words. Gray circles indicate edge crossings.

Although examples of real sentences with non-crossing dependencies are well-known (e.g., Mel'čuk 1988) the ungrammatical sentence in Fig. 2 has been chosen to introduce one of the problems that will be addressed in this article: what is a priori the maximum of number of crossings that can be reached? Crossings in syntactic dependency structures are rather rare (Havelka 2007) and it was hypothesized that this could be a side effect of minimizing the distance between syntactically linked words (Ferrer-i-Cancho 2006), which would be in turn a consequence of minimizing the online memory cost of the sentence (Morril 2000, Hawkins 2004, Grodner & Gibson 2005). Dependency lengths and crossings are no dissociated concepts as one may a priori believe (Hochberg & Stallmann 2003, Ferrer-i-Cancho 2006, Liu 2008).

This raises a very important research question for theoretical linguistics: is the minimization of crossings a principle by its own or is it a side-effect of a principle of dependency length minimization? Another related question is the origins of the low degree of vertices in syntactic dependency trees (in a sufficiently large sentence, vertices with a degree of the order of the length of the sentence are rare). In the sentence in Fig. 1, the maximum degree is three although it could be *n* - 1 = 8. Is it due to an autonomous principle of degree minimization or would it be again a side-effect of distance minimization? These questions are crucial for the development of a theory of language as simple as possible. A fundamental theoretical question is whether the low frequency of crossings or the low hubiness of syntactic dependency structures is due to an innate or biologically determined faculty for language that imposes universal constraints on world languages (e.g., the minimization of hubiness or the number of crossings) or these features could be simply due to the universal limitations of a complex brain for performing computations, being language production and processing particular cases of those computations (Christiansen et al 2012). Here it will be shown that the maximum number of crossings that can be achieved by a sentence ($C_{max}$) is bounded above by its mean dependency length ($\langle d \rangle$) and thus pressure for reducing crossings or hubiness could be a simple consequence of universal computational limitations of brains.

Another important research question is whether the properties of dependency structures, when considered independently of how vertices are arranged sequentially, exhibit features that help to save computational costs. Here it will be shown that the variance of vertex degrees determines the minimum $\langle d \rangle$ the can be achieved ($\langle d \rangle_{min}$), which in turn determines the minimum cognitive cost of sequences. This has a concrete consequence: the syntactic trees of long sentences cannot have hubs (hubs are vertices with a large number of links) due to the high online memory cost this would imply.

Those arguments are abstract enough to be valid not only for the communicative sequential behavior of other species but also for non-linguistic sequential behavior in general (human or not). In the present article, human language is the fuel to contribute to the development of a theory of natural sequential processing.

Besides illuminating the questions above, the present article aims at providing some mathematical results that are potentially useful for any research on (a) the mean dependency length (b) the number of crossings or (c) the relationship between mean dependency length and number of crossings in syntactic dependency trees. Lower and upper bounds for these quantities will be provided and the relationships between them will be unraveled.

The remainder of the article is organized as follows. Section 2 provides an introduction to graph theory that will help in the next sections. Sections 3 and 4 provide some results on dependency length and crossings, respectively. Sections 3 and 4 are essentially an enumeration of results aimed at facilitating their application. Readers interested in further details are referred to the appendices. The main article ends with a discussion in Section 5.

## 2. Graph theory

This section summarizes some results from standard graph theory and Appendix A. First we review elementary concepts of standard graph theory (Bollobás 1998). We neglect the direction of syntactic dependency arcs because our definition of dependency length and crossing is independent from it. A tree of *n* vertices has *n* - 1 edges. The degree of a vertex is the number of connections. For instance, 'she' in Fig. 1 has degree 1 while 'loved' has degree 3. Vertices with a large degree with regard to *n* are called hubs (Pastor-Satorras & Vespignani 2004) whereas vertices with degree one are called leaves (Bollobás 1998). It is convenient to label vertices not with the associated word (which is problematic if the same word appears more than once) but with natural numbers from 1 to *n*. Thus, $k_i$ is the vertex degree of the *i*-th word of the sentence (e.g. $k_1 = 1$, $k_2 = 3$ in Fig. 1). The structure of the tree is defined by the adjacency matrix $A = \{a_{ij}\}$, where $a_{ij} = 1$ if the pair of vertices $(i,j)$ is linked and otherwise $a_{ij} = 0$. The matrix is symmetric $a_{ij} = a_{ji}$ because we treat arcs as if they had no direction. Loops are not allowed ($a_{ii} = 0$). One has

$$k_i = \sum_{j=1}^{n} a_{ij} = \sum_{j=1}^{n} a_{ji}. \tag{1}$$

$\langle k \rangle$ and $\langle k^2 \rangle$ are the mean values of $k_i$ and $k_i^2$ (the 1st and 2nd moments of $k_i$, respectively), i.e.

$$\langle k \rangle = \frac{1}{n} \sum_{i=1}^{n} k_i. \tag{2}$$

$$\langle k^2 \rangle = \frac{1}{n} \sum_{i=1}^{n} k_i^2. \tag{3}$$

For any tree, it is easy to see that (Noy 1998)

$$\langle k \rangle = 2 - \frac{2}{n}. \tag{4}$$

for $n \geq 1$, knowing Eq. 2 and

$$\sum_{i=1}^{n} k_i = 2(n-1). \tag{5}$$

Since $\langle k \rangle$ is the same for any tree of a given length, $\langle k^2 \rangle$ determines *V*[*k*], the variance of the vertex degrees as $V[k] = \langle k^2 \rangle - \langle k \rangle^2$.

Two kinds of extreme trees that will be very useful throughout this article, i.e. the linear tree and the star tree, will be introduced next. A linear tree (also called path tree) is a tree with no branching at all (Fig. 1 (a)). A star tree is a tree where all vertices except one (the hub) are connected to the hub (Fig 3 (b)). Star trees model the syntactic dependency structure of utterances with a single head (the head being the hub). *V*[*k*] is maximized by star trees and thus $\langle k^2 \rangle$ alone can be regarded as a measure of "hubiness".

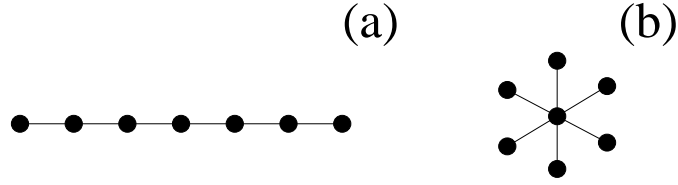

Figure 3. (a) a linear tree and (b) a star tree

Table 1 shows a summary of the second moment and the variance of linear and star trees (details of the calculation are given in Appendix A). It will be shown that $\langle k^2 \rangle$ is a key quantity for $d$ and $C$ that is maximized by star trees and minimized by linear trees. Table 2 shows some graph theoretic measurements on the dependency trees of Figs. 1 and 2.

Table 1
Summary of the properties of two extreme kinds of trees: star and linear trees. $n$ is the number of vertices, $\langle k^2 \rangle$ is the degree 2nd moment, $V[k]$ is the variance of the degree, $\langle d \rangle_{min}$ is the actual minimum value of $\langle d \rangle$ that a linear arrangement of vertices can achieve and $C$ is the number of link crossings

|  | Linear | Star |
|---|---|---|
| $\langle k^2 \rangle$ | $4 - \dfrac{6}{n}$ | $n - 1$ |
| $V[k]$ | $\dfrac{2}{n}\left(1 - \dfrac{2}{n}\right)$ | $n - 5 + \dfrac{4}{n}\left(2 - \dfrac{1}{n}\right)$ |
| $\langle d \rangle_{min}$ | 1 | $\dfrac{n^2}{4(n-1)}$ if $n$ is even $\dfrac{n+1}{4}$ if $n$ is odd |
| $C$ | $\leq \dfrac{n(n-5)}{2} + 3$ | 0 |

## 3. Length theory

This section summarizes results from Appendix B. $d_i$ is defined as the length of the $i$-th edge of dependency tree of $n$ vertices. $d_1,\ldots,d_i,\ldots,d_{n-1}$ is the list of the lengths of the $n$ - 1 edges of the tree. The mean dependency length of that tree is then

$$\langle d \rangle = \frac{1}{n-1} \sum_{i=1}^{n-1} d_i \qquad (6)$$

for $n \geq 1$. One has $\langle d \rangle = 11/8 \approx 1.375$ for the sentence in Fig. 1.

Table 2
Summary of the properties of the syntactic dependency trees of Fig. 1 and Fig. 2.

|  |  | Fig. 1 | Fig. 2 |
|---|---|---|---|
| Graph Theory | $n$ | 9 | |
|  | $\langle k^2 \rangle$ | 4 | |
| Length Theory | $\langle d \rangle$ | 11/8 = 1.375 | 29/8 = 3.625 |
|  | $\langle d^2 \rangle$ | 17/8 = 2.125 | 133/8 = 16.625 |
|  | $\langle d \rangle_{min}$ | $\geq 19/16 = 1.1875$ | |
|  | $E[d], E[\langle d \rangle]$ | $= 10/3 \approx 3.33$ | |
|  | $\langle d^2 \rangle$ | 17/8 = 2.125 | 133/8 = 16.625 |
| Crossing Theory | $C$ | 0 | 9 |
|  | $C_{max}$ (by degree, Eq. 14) | $\leq 18$ | |
|  | $C_{max}$ (by length, Eq. 12) | $\leq 3$ | $\leq 21$ |
|  | $C_{max}$ (by length, Eq. 13) | $\leq 9$ | $\leq 32$ |

We are interested in knowing the minimum and maximum values that $\langle d \rangle$ can take, $\langle d \rangle_{min}$ and $\langle d \rangle_{max}$, respectively. We would like to shed light on the extent to which actual sentences minimize or maximize $\langle d \rangle$. Since $1 \leq d_i \leq n - 1$, one has that $1 \leq \langle d \rangle \leq n - 1$. In general, 1 is the minimum value that $\langle d \rangle$ can take. This value is achieved by a linear tree whose vertices are arranged linearly. A linear tree is a tree where all vertices have degree 2 except two vertices that have degree 1. A linear arrangement of a linear tree consists of placing the vertices of degree 1 in both extremes the sequence (see Fig. 3 (a)) and placing the vertices of degree 2 immediately between its two linked vertices. Thus, $d_i = 1$ for all edges. While 1 is a reachable lower bound of $\langle d \rangle$ for linear trees, $n - 1$ is not a tight upper bound of $\langle d \rangle$ in general because there can only be a single edge of length $n - 1$. The number of edges that can be formed at distance $d$ is $N(d) = n - d$, hence $N(n - 1) = 1$.

A non-crossing tree is defined as linear arrangement of a tree without link crossings. The tree in Fig. 1 is non-crossing ($C=0$) while the tree in Fig. 2 is not ($C>0$). It can be shown that the maximum value of $\langle d \rangle$ that a non-crossing tree of $n$ vertices can achieve is

$$\langle d \rangle_{max} = \frac{n}{2} \tag{7}$$

with $\langle d_0 \rangle_{max} = \langle d \rangle_{max} - 1$.

As a star tree cannot have crossings because all vertices except the hub are connected to the hub, Eq. 7 gives the maximum value of $\langle d \rangle$ that a star tree can achieve. This maximum is achieved when the hub is placed first or last in the sequence of vertices. In contrast, the minimum value of $\langle d \rangle$ that a star tree can achieve is obtained when the hub is placed at the center and half of the leaves to its left and half of the leaves to its right (at position $(n + 1)/2$ if $n$ is odd and either at positions $n/2$ or $n/2 + 1$ if $n$ is even).

If the vertices of an edge are placed at random positions of a sentence (being a priori all the $n$ sentence positions equally likely), it can be can also be shown that the expected length of a single edge and its variance for $n \geq 2$ are

$$E[d] = \frac{n+1}{3} \qquad (8)$$

and

$$V[d] = \frac{(n+1)(n-2)}{18}, \qquad (9)$$

respectively. One has $E[d_0] = E[d] - 1$ and $V[d_0] = V[d]$. $E[\langle d \rangle]$, the expected mean length of the edges of a tree in which vertices have been placed at random, satisfies $E[\langle d \rangle] = E[d]$.

The minimum value that $\langle d \rangle$ can achieve is 1, which is only achieved by a linear tree. However, notice that $\langle d \rangle = 1$ is impossible to achieve in a tree with at least one vertex of degree three or greater. Hence, what about non-linear trees?

Table 1 shows the value of $\langle d \rangle_{min}$ for star trees. A lower bound for $\langle d \rangle_{min}$ can be derived from $\langle d \rangle_{min}$ for star trees. $\langle d \rangle_{min}$, the minimum value of $\langle d \rangle$, obeys

$$\langle d \rangle_{min} \geq \frac{1}{2(n-1)} \sum_{i=1}^{n} \left( \left\lfloor \frac{k_i}{2} \right\rfloor \left( \left\lfloor \frac{k_i}{2} \right\rfloor + 1 \right) + \frac{k_i + 1}{2} (k_i \bmod 2) \right), \qquad (10)$$

where $x \bmod y$ is the modulus of the division of $x$ by $y$. Eq. 10 is obtained by looking at the whole tree as an ensemble of star trees formed by each vertex and its neighbours (the star tree of the $i$-th vertex has $k_i+1$ vertices) and considering that every star tree is arranged sequentially in the best possible way, independently from other star trees. A much simpler lower bound for $\langle d \rangle_{min}$ with regard to Eq. 10 is

$$\langle d \rangle_{min} \geq \frac{n \langle k^2 \rangle}{8(n-1)} + \frac{1}{2}. \tag{11}$$

Eq. 11 shows that the minimum dependency length is bounded below by the variance of the degrees. Table 2 shows some dependency length measurements for the dependency trees of Figs. 1 and 2.

## 4. Crossing theory

This section summarizes results from Appendix C. Crossings are impossible ($C = 0$) for $n \leq 3$. When $n > 3$, simple upper bounds for $C_{max}$, the maximum number of crossings, are offered by the linear arrangement of vertices and by the structure of the tree. As for the former, one has

$$C_{max} \leq \binom{n-1-M}{2}, \tag{12}$$

where $M$ is the number of uncrossable edges (edges of length 1 or $n - 1$ are not crossable). Incorporating information from all dependency lengths, one also has

$$C_{max} \leq \frac{n-1}{2}\left(n\langle d \rangle - \langle d^2 \rangle - n + 1\right), \tag{13}$$

where $\langle d^2 \rangle$ is the 2$^{nd}$ moment of dependency length. It is easy to see from the previous inequality that crossings are impossible ($C = 0$) when $\langle d \rangle$ takes its absolute minimum value ($\langle d \rangle = 1$). Notice that Eq. 10 indicates that not all trees can reach $\langle d \rangle = 1$. As for an upper bound derived from the structure of the tree, one has

$$C_{max} \leq C_{pairs} = \frac{n}{2}\left(n - 1 - \langle k^2 \rangle\right), \tag{14}$$

where $C_{pairs}$ is the number of edge pairs that can cross (edges departing from the same vertex cannot cross).

Knowing that $\langle k^2 \rangle = n - 1$ in a star tree (Table 1), Eq. 14 gives that a star tree cannot have crossings ($C_{max} = 0$) regardless of how its vertices are arranged linearly. Since $C \geq 0$ it follows from Eq. 14 that a tree with $\langle k^2 \rangle > n - 1$ cannot exist because it would have a negative number of crossings. Therefore, a star tree has maximum $\langle k^2 \rangle$.

# 5. DISCUSSION

It has been shown that $\langle d \rangle_{min}$ is bounded below by $\langle k^2 \rangle$, i.e. the larger the value of $\langle k^2 \rangle$ (Eq. 11) the larger the value of $\langle d \rangle_{min}$. It has also been shown that $C_{max}$ is bounded above by both $\langle d \rangle$ (Eq. 13) and $\langle k^2 \rangle$ (Eq. 14), i.e. the smaller the value of $\langle d \rangle$ the smaller the value of $C_{max}$ while the larger the value of $\langle k^2 \rangle$ the smaller the value of $C_{max}$. This suggests that the low frequency of crossings in languages could be due to pressure for high degree variance but also to pressure for short dependency lengths. However, a high degree variance increases the minimum arc length that can be achieved and therefore raises the minimum cognitive cost of the sentence and thus the true reason for the low frequency of crossings in language might not hubiness but online memory limitations of the human brain.

Temperley (2008) has suggested that the structural properties of syntactic dependency trees (leaving aside the linear arrangement of vertices) might reflect pressure for dependency length minimization. With this regard, our results have implications for the presence of hubs in sentences. Eq. 14 implies that the more skewed the degree distribution of vertices (the higher the value of $\langle k^2 \rangle$), the higher the minimum value of $\langle d \rangle$ that can be achieved. Reading this result in terms of the cognitive cost implied by $\langle d \rangle$ (Ferrer-i-Cancho 2006), long sentences with large $\langle k^2 \rangle$ would be cognitively too expensive in practice. If actual sentences minimize $\langle d \rangle$, then a necessary condition is that $\langle d \rangle_{min}$ is not too high. Thus, $\langle k^2 \rangle$ must be reduced and hubs must be avoided. This is in contrast with the large-scale organization of syntactic dependency networks (Ferrer-i-Cancho et al. 2004), where vertices with high degree do exist. The absence of hubs at the sentence scale is likely to be caused by the constraints of short term memory (Morrill 2000, Hawkins 2004, Grodner and Gibson 2005) while the existence of hubs at the large-scale could be due to the fact that dependencies at this scale are kept by long-term memory. In sum, the limited resources of our brains lead to the principle of dependency length minimization (Morrill 2000, Hawkins 2004, Grodner and Gibson 2005, Ferrer-i-Cancho 2006), which in turn make hubs expensive in syntactic dependency trees.

Our theoretical framework suggests new questions for empirical research. If there is actually cognitive pressure for reducing hubiness (V[$k$]) or mean arc lengths ($\langle d \rangle$), an important research question is: how do these quantities scale with $n$, the length of the sentence? As the maximum number of crossings depends on V[$k$] or $\langle d \rangle$ (Section 3), how does $C$ scales as a function of V[$k$] or $\langle d \rangle$? As the minimum value of $\langle d \rangle$ depends on V[$k$] (Section 2), how does $\langle d \rangle$ scale as a function of V[$k$]? The growing availability of dependency treebanks (e.g. Civit *et al.* 2006, Böhmová *et al.* 2003, Bosco *et al.* 2000) suggests that the questions above could be answered for syntactic dependency trees in a near future.

Our results have also implications for the parallel research on complex network physics. It has been shown that $\langle k^2 \rangle$ is a crucial quantity for $\langle d \rangle_{min}$ (Eq. 11), $C_{max}$ (Eq. 14) in dependency trees. This result is reminiscent of the key role played by $\langle k^2 \rangle / \langle k \rangle$ in large complex networks (Pastor-Satorras & Vespignani 2004), for instance, concerning the diffusion of epidemics in Internet (if $\langle k^2 \rangle / \langle k \rangle$ diverges then the pandemics cannot be stopped). In syntactic dependency trees, one has that $\langle k^2 \rangle / \langle k \rangle = \langle k^2 \rangle / (2 - 2/n)$). Our findings support the idea that $\langle k^2 \rangle / \langle k \rangle$ is a general fundamental property of the network of many real systems.

ACKNOWLEDGEMENTS

This manuscript derives from a longer manuscript finished in 2007. We are grateful to G. Altmann, O. Jiménez and H. Liu for helpful comments.. This work was supported by the grant "Iniciació i reincorporació a la recerca" from the Universitat Politècnica de Catalunya and the grants BASMATI (TIN2011-27479-C04-03) and OpenMT-2 (TIN2009-14675-C03) from the Spanish Ministry of Science and Innovation.

# APPENDIX A: GRAPH THEORY

### A.1. 2$^{nd}$ moment and variance of degree in linear and star trees

Knowing Eq. 3, it is easy to see that a linear graph (i.e. two vertices of degree 1 and the remainder of degree 2) has

$$\langle k^2 \rangle = \frac{1}{n}(2 + 4(n-2)) = 4 - \frac{6}{n} \tag{A1}$$

whereas a star graph has

$$\langle k^2 \rangle = \frac{1}{n}(n-1 + (n-1)^2) = n-1 \tag{A2}$$

for $n \geq 2$. While $\langle k^2 \rangle$ never exceeds 4 in a linear graph it grows linearly with *n* in a star graph. Knowing that the degree variance is $V[k] = \langle k^2 \rangle - \langle k \rangle^2$ and Eqs. 4, A1 and A2, it is easy to show that a linear graph has

$$V[k] = \frac{2}{n}\left(1 - \frac{2}{n}\right). \tag{A3}$$

and a star graph has

$$V[k] = n - 5 + \frac{4}{n}\left(2 - \frac{1}{n}\right). \tag{A4}$$

See Noy (1998) for $\langle k^2 \rangle$ and $V[k]$ in random trees and random trees without crossings.

**A.2. Linear trees have minimum degree variance.**

Next it will be proven that a linear tree has minimum $\langle k^2 \rangle$ by induction on $n$. Consider the sum of the squares of degrees of a tree of $n$ vertices is

$$K_2(n) = \sum_{i=1}^{n} k_i^2 \tag{A5}$$

and thus $\langle k^2 \rangle = K_2(n)/n$. In a linear, tree Eq. A1 gives $K_2(n) = 4n - 6$. We want to prove that

$$K_2(n) \geq 4n - 6 \tag{A6}$$

for any tree (with $n \geq 2$). When $n = 2$, Eq. 6 holds trivially as only a linear tree is possible. We hypothesize that A6 holds for $n$ and wonder it holds for $n + 1$, too. Imagine that the degree sequence of a tree of $n + 1$ vertices is $k_1, k_2, k_3,\ldots, k_n, k_{n+1}$. A leaf is defined as a vertex of degree 1. It is well-known that any tree has at least two leaves (Bollobás 1998, pp. 11). Without any loss of generality, consider that the $(n+1)$-th vertex is a leaf and that the vertex that must be attached to that leaf is the $n$-th vertex (a leaf, by definition, has one connection). As $k_{n+1} = 1$, the tree of $n+1$ vertices has

$$K_2(n+1) = \sum_{i=1}^{n+1} k_i^2 = \sum_{i=1}^{n} k_i^2 + 1. \tag{A7}$$

The degree sequence $k_1, k_2, k_3,\ldots, (k_n - 1)$ defines a tree of $n$ vertices as we only have substracted a leaf. As $k_n^2 = (k_n - 1)^2 + 2k_n - 1$, Eq. A7 can be rewritten as

$$K_2(n+1) = \sum_{i=1}^{n-1} k_i^2 + (k_n - 1)^2 + 2k_n = K_2'(n) + 2k_n, \tag{A8}$$

where $K_2'(n)$ is the value of $K_2(n)$ for the degree sequence of length $n$ above. By the hypothesis of induction, $K_2'(n) \geq 4n - 6$ and thus

$$K_2(n+1) \geq 4n - 6 + 2k_n. \tag{A9}$$

Notice that $k_n \geq 1$ as the $n$-th vertex is connected to the $(n+1)$-th vertex. Furthermore, notice also that $k_n \geq 2$ when $n > 2$ because the $n$-th vertex must be connected to vertices other than the $(n + 1)$-th to keep the graph connected (connectedness of the graph of $n+1$ nodes requires $k_n > 1$ except when $n = 1$, but we are considering the case $n > 2$). Applying $k_n \geq 2$ to Eq. A9 yields

$$K_2(n+1) \geq 4n - 2 = 4(n+1) - 6 \tag{A10}$$

as we wanted to prove.

# APPENDIX B: LENGTH THEORY

**B.1. The distribution of dependency lengths in random linear arrangements.**

First we study the distribution of dependency lengths in trees where vertices are placed at random in a sequence. The probability that two randomly placed vertices in a sequence of length *n* are at distance *d* is (Ferrer-i-Cancho 2004)

$$p(d) = \frac{N(d)}{\sum_{i=1}^{n-1} N(i)}, \tag{B1}$$

where $N(d) = n - d$ is the number of vertex pairs at distance *d* ($N(d) = 0$ if $d < 1$ or $d > n - 1$). Knowing Table 3 and $N(d) = n - d$, Eq. B1 is transformed into

$$p(d) = \frac{2(n-d)}{n(n-1)}. \tag{B2}$$

for $n \geq 2$. $p(d)$ also defines the probability that the vertices forming an edge are at distance *d* (independently from the length of other edges). Thus, $E[d]$, the expected value of the distance *d* separating two linked vertices is

$$E[d] = \sum_{d=1}^{n-1} p(d)d. \tag{B3}$$

Table 3
A summary of summations of powers of consecutive natural numbers
(Spiegel & Liu 1999)

| $a$ | $\sum_{x=1}^{n} x^a$ | $\sum_{x=1}^{n-1} x^a$ |
|---|---|---|
| 1 | $\dfrac{n(n+1)}{2}$ | $\dfrac{n(n-1)}{2}$ |
| 2 | $\dfrac{n(n+1)(2n+1)}{6}$ | $\dfrac{(n-1)n(2n-1)}{6}$ |
| 3 | $\dfrac{n^2(n+1)^2}{4}$ | $\dfrac{(n-1)^2 n^2}{4}$ |

Applying Eq. B2 and Table 3 to Eq. B3, it is obtained

$$E[d] = \frac{2}{n(n-1)}\left(n\sum_{d=1}^{n-1} d - \sum_{d=1}^{n-1} d^2\right) = \frac{n+1}{3}. \tag{B4}$$

for $n \geq 2$ after some algebra. Notice that $E[d]$ (Eq. B4) is the expected length of a single edge. $E[\langle d \rangle]$ is the expected mean arc length over all the edges of a tree (in which vertices have been randomly placed). It is easy to see that $E[d] = E[\langle d \rangle]$ for any tree because the expectation of a sum of random variables (independent or not) is the sum of the expectations of each of the variables (DeGroot 1989). Recalling the definition of $\langle d \rangle$ in Eq. 6, one has

$$E[<d>] = E\left[\frac{1}{n-1}\sum_{i=1}^{n-1} d_i\right] = \frac{1}{n-1}\sum_{i=1}^{n-1} E[d_i] = E[d]. \tag{B5}$$

as we wanted to prove.

$V[d]$, the variance of $d$ of a single edge, is

$$V[d] = E[d^2] - E[d]^2. \tag{B6}$$

Firstly, we calculate $E[d^2]$. Applying Eqs. B2 and B3 to

$$E[d^2] = \sum_{d=1}^{n-1} p(d) d^2, \tag{B7}$$

it is obtained

$$E[d^2] = \frac{2}{n(n-1)}\left(n\sum_{d=1}^{n-1} d^2 - \sum_{d=1}^{n-1} d^3\right). \tag{B8}$$

The application of Table 3 yields finally

$$E[d^2] = \frac{n(n+1)}{6} \tag{B9}$$

for $n \geq 2$ after some algebra.

Secondly, replacing the r.h.s. of Eqs. B4 and B9 into Eq. B6 one finally obtains

$$V[d] = \frac{(n+1)(n-2)}{18}, \tag{B10}$$

with $n \geq 2$ after some work.

As for $E[d_0]$, $E[d_0^2]$ and $V[d_0]$, knowing that $E[x - 1] = x$ and $V[x - 1] = V[x]$ (DeGroot 1989) and $d_0 = d - 1$, one obtains

$$E[d_0] = \frac{n-2}{3}, \tag{B11}$$

and $E[d_0^2] = E[d^2] - 2E[d] + 1 = n^2/6 + n/2 + 1/3$ and $V[d_0] = V[d]$. Eqs. B10 and B11 have also been derived in the context of the distance between not necessarily consecutive repeats in a sequence (Zörnig 1984).

**B.2. The maximum mean dependency length.**

We aim to calculate or bound above $\langle d \rangle_{max}$, the maximum value that $\langle d \rangle$ can reach in a linear arrangement of a tree without crossings. Two procedures to arrange the vertices linearly will be presented: one for star trees and another for linear trees. Then it will be shown that value of $\langle d \rangle$ achieved by those procedures is actually maximum.

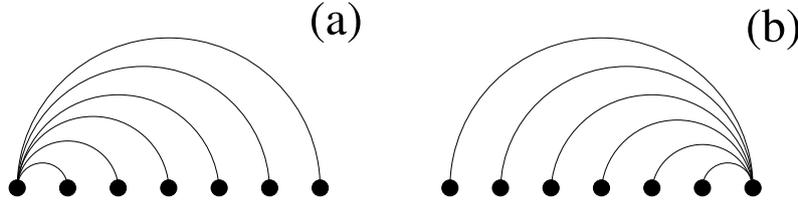

Figure 6. Two symmetric ways of arranging the vertices of a star tree in a way that the mean dependency length is $\langle d \rangle = n/2$.

Imagine that the hub of a star tree is placed at one of the extremes of the sequence of vertices (the hub is placed first or last) as in Fig. 6. In that case, the mean dependency length is

$$\langle d \rangle = \frac{1}{n-1} \sum_{i=1}^{n-1} d_i = \frac{1}{n-1} \sum_{i=1}^{n-1} d \; . \tag{B12}$$

Knowing Table 3, Eq. B12 yields

$$\langle d \rangle = \frac{n}{2} \tag{B13}$$

and

$$\langle d_0 \rangle = \langle d \rangle - 1 = \frac{n}{2} - 1 \; . \tag{B14}$$

It is tempting to think that star trees are the only trees that can achieve this mean dependency length. Indeed, it easy to see that linear trees arranged linearly as in Fig. 7 also achieve the same mean dependency length than star trees with hub first or last as those arrangements of linear trees also obey Eq. B12.

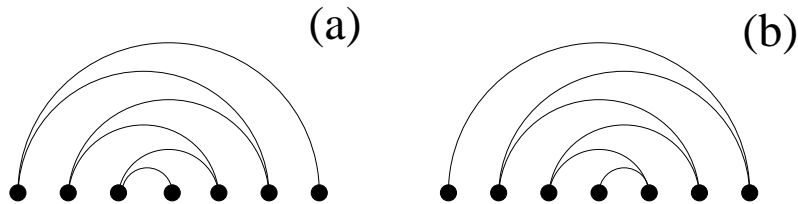

Figure 7. Two symmetric ways of arranging the vertices of a linear tree in a way that the mean dependency length is $\langle d \rangle = n/2$.

$D$ is defined as the sum of dependency lengths, i.e. $D = (n-1)\langle d \rangle$ and $\Delta(x) = x(x-1)/2$. Next it will be shown by induction on $n$ that a non-crossing tree with $D = \Delta(n)$ (and thus $\langle d \rangle = n/2$) has the maximum $D$ that a non-crossing tree can achieve. The base of the induction is $n = 2$, where only a non-crossing tree can be formed. In that case $D = 1$ is maximum. The induction hypothesis is that any non-crossing tree of $n' < n$ vertices with $D = \Delta(n')$ has maximum $D$. It will be shown that a non-crossing tree of $n$ vertices ($n \geq 3$) and $D = \Delta(n)$ also has maximum $D$. To see it, consider that any non-crossing tree of $n$ vertices can be constructed in two ways (Yuret 2006):

a) Concatenating two non-crossing subtrees that share the $v$-th vertex of the sequence (Fig. 8 (a)). That vertex is the last vertex of the first subtree and the first vertex of the second subtree. One subtree has $v$ vertices and the other subtree has $n-v+1$ vertices. $2 \leq v \leq n - 1$ is required for being a true decomposition of a non-crossing tree of $n$ vertices (each subtree having less than $n$ vertices). For instance, the tree in Fig. 1 can be constructed by concatenating the subtree induced by words from 'She' to 'for' (both included) and the one induced by words from 'for' to 'passed' (both included).

b) Concatenating two non-crossing subtrees that do not share any vertex, one with $v$ vertices and the other with the following $n-v$ vertices, and linking the first vertex of the first subtree with the last vertex of the second subtree (Fig. 8 (b)). $1 \leq v \leq n - 1$ is required for being a decomposition of a non-crossing tree of $n$ vertices. The non-crossing tree in Fig. 1 has not been constructed in this fashion but this is the case of the subtree induced by the words 'for', 'the' and 'dangers'.

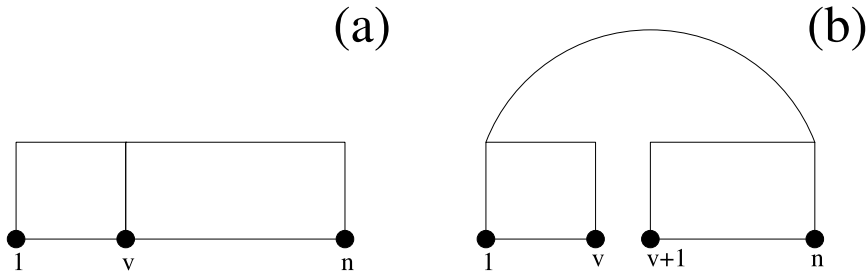

Figure 8. Schemes of two decompositions of a non-crossing tree. Rectangles indicate non-crossing subtrees. Circles indicate the first and the last vertex of each rectangle. In (a), the last vertex of the first subtree and the first vertex of the second subtree overlap. In (b), the subtrees are joined by a link between the first vertex of the first subtree and the last vertex of the second subtree.

$D_a(v)$ and $D_b(v)$ are defined as the maximum sum of arc lengths for construction a) and b), repectively, as a function of $v$, the position of the last vertex of the first non-crossing subtree. As for construction of type a), the maximum sum of dependency lengths that can be reached is

$$D_a = \max_{2 \leq v \leq n-1} (D_a(v)). \qquad (B15)$$

By the hypothesis of induction, $D_a(v)$ is

$$D_a(v) = \Delta(v) + \Delta(n - v + 1) = v^2 - (n+1)v + \frac{n(n+1)}{2}. \qquad (B16)$$

As for constructions of type b), the maximum sum of dependency lengths that can be reached is

$$D_b = \max_{1 \leq v \leq n-1}(D_b(v)).\qquad(B17)$$

By the hypothesis of induction, $D_b(v)$

$$D_b(v) = n - 1 + \Delta(v) + \Delta(n-v) = v^2 - nv + \frac{n(n+1)}{2} - 1.\qquad(B18)$$

If is easy to show that construction a) produces smaller sums of arc lengths than construction b) because

$$D_b(v) = D_a(v) + v - 1.\qquad(B19)$$

for $2 \leq v \leq n-1$ and then $D_b(v) > D_a(v)$ within that range of $v$.

Using $dD_b(v)/dv = 2v - n = 0$ it is easy to see that $D_b(v)$ has only one critical point within the interval $[1, n-1]$, i.e. $v = n/2$. As $d^2D_b(v)/dv = 2 > 0$, $D_b(v)$ has a minimum at $v = n/2$ and therefore $D_b(1)$ and $D_b(n-1)$ are equal maxima within that interval (by symmetry, $D_b(1) = D_b(n-1)$, recall Eq. B18). Therefore the maximum $D$ is

$$D_b(1) = D_b(n-1) = \frac{n(n-1)}{2} = \Delta(n)\qquad(B20)$$

as we wanted to prove.

**B.3. The minimum mean dependency length.**

We aim to find a lower bound for $\langle d \rangle$ given the degree of each vertex. $\tau_i$ is defined as the sum of the lengths of the links formed with the *i*-th vertex. $\langle d \rangle$ can be written in terms of $\tau_i$, i.e.

$$\langle d \rangle = \frac{1}{2(n-1)}\sum_{i=1}^{n}\tau_i.\qquad(B21)$$

$k_i$ is defined as the degree of the *i*-th vertex. We aim to find the minimum value of $\tau_i$. This is equivalent to finding the minimum value of $\langle d \rangle$ for the star tree of $n = k_i + 1$ vertices defined by the *i*-th vertex and its $k_i$ adjacent vertices (notice $\langle d \rangle = \tau_i/(k_i + 1)$ in that case).

If $k_i$ is an even number, the minimum $\tau_i$ is obtained by placing $k_i/2$ of the adjacent vertices immediately before vertex *i* and $k_i/2$ of the remaining vertices immediately after, that is,

$$\tau_i \geq 2\sum_{j=1}^{\frac{k_i}{2}} j.\qquad(B22)$$

If $k_i$ is an odd number, the minimum $\tau_i$ is obtained by placing $k_i/2+1$ of the adjacent vertices immediately before vertex $i$ and $k_i/2$ of the remaining adjacent vertices immediately after it or by the symmetric configuration (i.e. placing $k_i/2$ of the adjacent vertices immediately after vertex $i$ and $k_i/2+1$ of the remaining adjacent vertices immediately after it). Therefore,

$$\tau_i \geq 2 \sum_{j=1}^{\frac{k_i-1}{2}} j + \frac{k_i+1}{2}. \tag{B23}$$

Merging Eqs. B22 and B23, one obtains

$$\tau_i \geq 2 \sum_{j=1}^{\left\lfloor \frac{k_i}{2} \right\rfloor} j + \frac{k_i+1}{2}(k_i \bmod 2), \tag{B24}$$

being $x \bmod y$ is the modulus of the division of $x$ by $y$.

It is easy to see that this kind of arrangement of adjacent vertices around the $i$-th vertex is optimal (minimizes $\tau_i$). If the $i$-th vertex is placed at position $\pi$, the nearest placements for an adjacent vertex are either positions $\pi - 1$ or $\pi+1$. If these two positions are already taken by adjacent vertices, the nearest positions available are $\pi-2$ and $\pi+2$, and so on.

Replacing Eq. B24 into Eq. B21, one gets

$$\langle d \rangle_{min} \geq \frac{1}{2(n-1)} \sum_{i=1}^{n} \left( \left\lfloor \frac{k_i}{2} \right\rfloor \left( \left\lfloor \frac{k_i}{2} \right\rfloor + 1 \right) + \frac{k_i+1}{2}(k_i \bmod 2) \right). \tag{B25}$$

A lower bound of $<d>_{min}$ that is simpler than that of Eq. B25 can be obtained. When $k_i$ is even, Eq. B22 is equivalent to

$$\tau_i \geq \frac{k_i}{2}\left(\frac{k_i}{2}+1\right) = \frac{k_i^2}{4} + \frac{k_i}{2}. \tag{B26}$$

When $k_i$ is odd. Eq. B23 is equivalent to

$$\tau_i \geq \left(\frac{k_i+1}{2}\right)^2 = \frac{k_i^2}{4} + \frac{k_i}{2} + \frac{1}{4}. \tag{B27}$$

Regardless of whether $k_i$ is even or not, $\tau_i$ is bounded below by Eq. B26 and then Eq. B21 becomes

$$\langle d \rangle_{min} \geq \frac{1}{2(n-1)} \sum_{i=1}^{n} \left( \frac{k_i^2}{4} + \frac{k_i}{2} \right). \tag{B28}$$

After some algebra, one obtains

$$\langle d \rangle_{min} \geq \frac{n}{4(n-1)} \left( \frac{\langle k^2 \rangle}{2} + \langle k \rangle \right). \tag{B29}$$

Replacing $\langle k \rangle = 2 - 2/n$ (Eq. 4), into Eq. B29 it is obtained finally

$$\langle d \rangle_{min} \geq \frac{n \langle k^2 \rangle}{8(n-1)} + \frac{1}{2}. \tag{B30}$$

If we consider a linear tree, there are $n-2$ vertices where $k_i = 2$ and 2 vertices where $k_i = 1$, so Eq. B25 gives $\langle d \rangle_{min} = 1$, which is indeed the actual minimum for this kind of tree. We could also consider a star tree, where all vertices have $k_i = 1$ except the hub, which has $k_i = n - 1$. It is tempting to use Eq. B25 to bound $\langle d \rangle_{min}$ below but the contribution of vertices of degree 1 will be underestimated. For this reason, it is convenient to consider

$$\langle d \rangle_{min} = \tau_h/(n-1), \tag{B31}$$

where $\tau_h$ is the true minimum value of $\tau_i$ that the hub can achieve. Eqs. B26 and B27 indicate that

$$\tau_h = \frac{n^2}{4}. \tag{B32}$$

if $n$ is even (the hub has ood degree) and

$$\tau_h = \frac{n-1}{2} \left( \frac{n-1}{2} + 1 \right). \tag{B33}$$

if $n$ is odd (the hub has even degree). Applying Eqs. B32 and B33 to Eq. B31, it is obtained that a star tree has

$$\langle d \rangle_{min} = \frac{n^2}{4(n-1)} \tag{B34}$$

if $n$ is even and

$$\langle d \rangle_{min} = \frac{n+1}{4} \tag{B35}$$

if $n$ is odd.

## APPENDIX C: CROSSING THEORY

We aim to bound above $C$, the number of link crossings. $C=0$ for $n \leq 3$ (if $n \leq 2$, the number of edges does not exceed 1 and thus crossings are impossible; if $n=3$, the two edges cannot cross as they have a vertex in common). Hereafter, $n>3$ is assumed. We do not aim to calculate $C_{max}$, the actual maximum number of crossings that a sentence can reach, but upper bounds of $C_{max}$.

**C.1. A simple upper bound for the number of crossings.**

If a sentence has $n$ vertices, then $C_{max}$ cannot exceed the number of different pairs of edges, i.e.

$$C_{max} \leq \binom{n-1}{2} = \frac{(n-1)(n-2)}{2} \tag{C1}$$

for $n \geq 2$.

**C.2. Upper bounds of the number of crossings from dependency lengths.**

Since no crossing can be formed with edges of length 1 or $n - 1$, the actual number of edges that can be involved in a crossing is $n - 1 - N_e(1) - N_e(n - 1)$ where $N_e(d)$ here is the actual number of edges whose length is $d$. Thus,

$$C_{max} \leq \binom{n-1-N_e(1)-N_e(n-1)}{2}. \tag{C2}$$

Configurations where crossings are impossible can be derived imposing that the number of edges that can cross is at most 1, i.e.

$$n - 1 - N_e(1) - N_e(n-1) \leq 1, \tag{C3}$$

which means that crossings are impossible if (a) there is no arc of maximum length ($N_e(n - 1) = 0$) and at most one arc has a length different than 1 ($n - 2 \leq N_e(1) \leq n - 1$) or (b) there is an arc of maximum length ($N_e(n - 1) = 1$) and at most one arc with a length between 1 and $n - 1$ ($n - 3 \leq N_e(1) \leq n - 2$).

Upper bounds of $C_{max}$ can be derived involving the length of each arc. Knowing that $d - 1$ is the number of vertices under an arc and $n - d - 1$ is the number of vertices "off the arc", the number of crossings with different arcs in which an arc of length $d$ can be involved cannot exceed $c(d) = (d-1)(n-d-1)$. Notice that $c(d)$ could exceed $n - 2$, the maximum number of crossings in which an arc can be involved (e.g., take $d=3$ and $n > 2$), but $c(d)$ is exact when $d = 1$ or $d = n - 1$ ($c(d)=0$ in both cases). If $d_i$ is the length of the $i$-th arc, one can write

$$C_{max} \leq \frac{1}{2} \sum_{i=1}^{n-1} c(d_i), \tag{C4}$$

which applying $c(d) = nd - d^2 - n + 1$ becomes

$$C_{max} \leq \frac{1}{2}\left(n\sum_{i=1}^{n-1} d_i - \sum_{i=1}^{n-1} d_i^2 - (n-1)^2\right) \tag{C5}$$

and finally

$$C_{max} \leq \frac{n-1}{2}\left(n\langle d\rangle - \langle d^2\rangle - n + 1\right). \tag{C6}$$

**C.3. Upper bounds of the number of crossings from vertex degrees.**

Upper bounds for $C_{max}$ based on the structure of the tree will be derived next. It is convenient to write $C$ as a function of the adjacency matrix $A = \{a_{ij}\}$,

$$C = \frac{1}{4}\sum_{i=1}^{n}\sum_{j=1}^{n}a_{ij}C_{ij}, \tag{C7}$$

where $C_{ij}$ is the number of crossings in which the pair of vertices $(i,j)$ is involved ($C_{ij}=0$ if $a_{ij}=0$). Notice that the definition of link crossing given in Section 1 implies that an edge connecting the pair of vertices $(i,j)$ cannot cross any edge formed with either $i$ or $j$ (including the edge under consideration itself). Thus the edge formed by the pair of vertices $(i,j)$ cannot cross any of the $k_i + k_j - 1$ edges (being $k_i$ the degree of the $i$-th vertex) formed involving vertex $i$ or vertex $j$. The number of edges that can be crossed by the edge formed by $(i,j)$ is thus $(n-1) - (k_i + k_j - 1) = n - k_i - k_j$. Thus, $C_{ij} \leq n - k_i - k_j$. $C_{pairs}$ is defined as the number of different edge pairs that can cross. Replacing $C_{ij}$ by its upper bound, i.e., $n - k_i - k_j$, in Eq. C7, it is obtained

$$C \leq C_{pairs} = \frac{1}{4}\sum_{i=1}^{n}\sum_{j=1}^{n}a_{ij}(n - k_i - k_j). \tag{C8}$$

The previous Eq. gives after some work

$$C_{pairs} = \frac{1}{2}\left(n(n-1) - \sum_{i=1}^{n}k_i^2\right) \tag{C9}$$

and finally

$$C_{pairs} = \frac{n}{2}\left(n - 1 - \langle k^2\rangle\right). \tag{C10}$$

Knowing that $\langle k^2\rangle = n - 1$ in a star graph, Eq. C10 means that a star graph cannot have crossings ($C=0$) regardless of how its vertices are arranged linearly as $0 \leq C \leq C_{pairs} \leq 0$ in that case. A linear tree, which has minimum $\langle k^2\rangle$ (Appendix A), transforms Eq. C10 with $\langle k^2\rangle = 4 - 6/n$ into

$$C_{pairs} = \frac{n(n-5)}{2} + 3. \tag{C11}$$